\documentclass[letterpaper, 10 pt, conference]{ieeeconf}  
\IEEEoverridecommandlockouts                                      
\usepackage{amsmath}
\usepackage{amssymb}
\usepackage{mathtools}
\usepackage{graphicx}
\usepackage{float}
\usepackage{array}
\usepackage{tikz}
\usepackage{listings}
\usepackage{hyperref}
\usepackage{graphicx}
\usepackage{cite}
\usepackage{dsfont}
\usepackage{mathtools,etoolbox}
\usepackage[none]{hyphenat}
\usepackage[utf8]{inputenc}
\usepackage[english]{babel}

\usepackage{amsthm}
\usepackage{xfrac}
\usepackage{nicefrac}
\usepackage{booktabs}
\usepackage[switch, pagewise]{lineno}
\usepackage{nicefrac}
\usepackage{flushend}
\usepackage{xcolor}
\usepackage{todonotes}

\usepackage{caption}

\hypersetup{colorlinks=true,urlcolor=blue,citecolor=red}

\usepackage{colortbl}
\usepackage{multirow}
\usepackage{color, colortbl}
\definecolor{LightCyan}{rgb}{0.88,1,1}

\makeatletter

\usepackage{algpseudocode}
\usepackage{algorithm}

\long\def\invis#1{}
\usepackage[some, placement=top]{background}

\backgroundsetup{
        placement=top,
        position={8cm,2cm}}

\SetBgScale{1}
\SetBgContents{\parbox{16cm}{\Large \centering This paper has been accepted for publication at the IEEE International Conference on Robotics and Automation (ICRA), Atlanta, 2025. \textcopyright IEEE}}
\SetBgColor{gray}
\SetBgAngle{0}
\SetBgOpacity{0.9}
\title{\LARGE \bf \textit{ODYSSEE}: Oyster Detection Yielded by Sensor Systems \\on Edge Electronics}
\author{Xiaomin Lin$^{1}$*\textsuperscript{\textdagger}, Vivek Mange$^{2}$*,  Arjun Suresh$^{1}$*, Bernhard Neuberger$^{3}$,  Aadi Palnitkar$^{1}$, \\ 
Brendan Campbell$^{4}$, Alan Williams$^{5}$,  Kleio Baxevani$^{2}$, 
Jeremy Mallette$^{8}$, Alhim Vera$^{6}$ \\Markus Vincze$^{3}$, Ioannis Rekleitis$^{2}$, Herbert G. Tanner$^{2}$, Yiannis Aloimonos$^{1}$ 
\thanks{* Equal Contributors, \textsuperscript{\textdagger} Corresponding Authors.}
\thanks{$^{1}$Maryland Robotics Center, University of Maryland, College Park, MD 20742, USA. Emails: \texttt{\{xlin01, arjsur, apalnitk, jyaloimo\}@umd.edu}.
        }%
\thanks{$^{2}$Center for Autonomous and Robotic Systems, University of Delaware, Newark, DE, 19711, USA. Emails: \texttt{\{vivekm, kleiobax, yiannirs,btanner\}@udel.edu.
        }}%
\thanks{$^{3}$ Automation and Control Institute, TU Wien, 1040 Vienna, Austria. \{neuberger, vincze\}@acin.tuwien.ac.at
}%
\thanks{$^{4}$ School of Marine Science and Policy, University of Delaware, Lewes, DE, 19958, USA. Email: \texttt{bpc@udel.edu}.
}%
\thanks{$^{5}$ University of Maryland Center for Environmental Science, Horn Point Laboratory, Cambridge, MD 21613, USA Email: \texttt{awilliams@umces.edu}.
}%
\thanks{$^{6}$ College of Engineering and Applied Science, University of Cincinnati, OH 45221, USA. Email: \texttt{veragoaa@mail.uc.edu}.
}%
\thanks{$^{8}$Indepedendent Robotics, Montreal, QC, Canada. Email: \texttt{jeremy.mallette@independentrobotics.com}.
}}

\begin{document}

\maketitle
                                                                                                                                                            \thispagestyle{empty}
\pagestyle{empty}


\begin{abstract}

Oysters are a vital keystone species in coastal ecosystems, providing significant economic, environmental, and cultural benefits. As the importance of oysters grows, so does the relevance of autonomous systems for their detection and monitoring. However, current monitoring strategies often rely on destructive methods. While manual identification of oysters from video footage is non-destructive, it is time-consuming, requires expert input, and is further complicated by the challenges of the underwater environment.

To address these challenges, we propose a novel pipeline using stable diffusion to augment a collected real dataset with photorealistic synthetic data. This method enhances the dataset used to train a YOLOv10-based vision model. The model is then deployed and tested on an edge platform; Aqua2, an Autonomous Underwater Vehicle (AUV), achieving a state-of-the-art 0.657 mAP@50 for oyster detection.
\end{abstract}

\section{INTRODUCTION}
\label{section:introduction}
\begin{figure}
 \centering
{\includegraphics[width=1.0\linewidth]{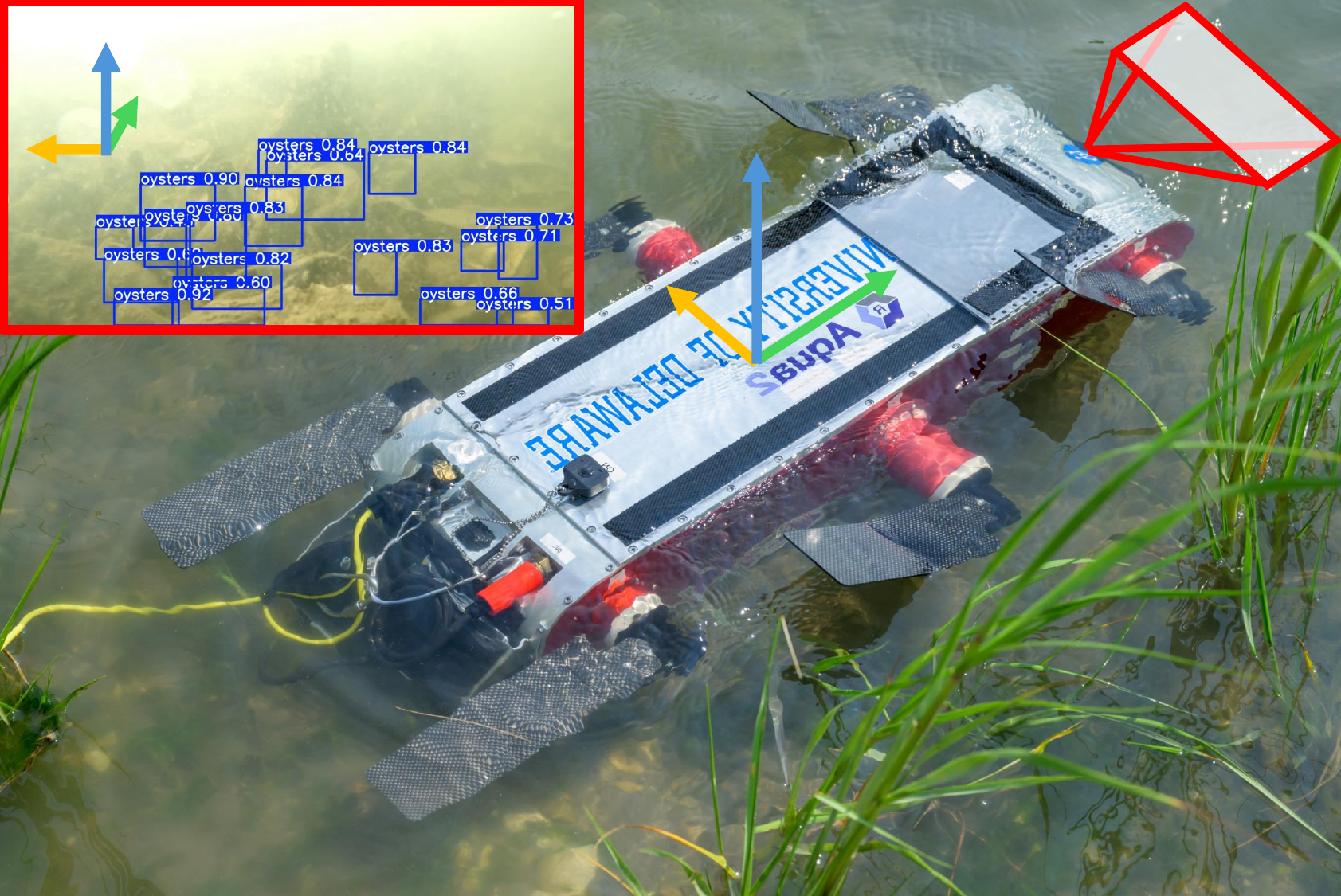}}
\captionsetup{font={footnotesize},labelfont=bf}
\caption{Oyster detection system deployed on an Autonomous Underwater Vehicle (AUV). The main image shows oyster detection in real-time in a shallow marine environment on the Aqua2 robot. The inset highlights the output bounding boxes around the oysters detected in situ}
\label{fig:detection_result}
\vspace{-5mm}
\end{figure}

Oyster reefs are crucial benthic environments, providing numerous ecosystem services, including water filtration, enhanced species richness, shoreline stabilization, and velocity attenuation~\cite{coen2007ecosystem,grabowski2007restoring,campbell2024assessing}. Unfortunately, standing oyster stocks in areas such as the Chesapeake Bay~\cite{newell1988ecological} and the North Sea~\cite{pogoda2019current} have decreased significantly 
due to over-fishing, global warming, and disease. 
To mitigate this devastating ecological loss, massive efforts are being made to monitor and restore oyster habitats across Europe~\cite{pogoda2019current}, Aisa~\cite{quan2017long,bishop2023oyster} and the United States~\cite{baggett2015guidelines, blomberg2018habitat, mcfarland2018restoring, theuerkauf2019integrating}.

Despite this effort, efficiently monitoring oyster beds for guided and enhanced restoration remains a major challenge~\cite{pine2022adaptive, fitzsimons2020restoring,baggett2015guidelines}. Traditional monitoring methods, such as those conducted annually in Delaware Bay, New Jersey~\cite{saw2024}, and biannually in Chesapeake Bay, Maryland~\cite{mdosau2020}, rely heavily on destructive sampling techniques (dredging or tonging) and trained technicians to
process samples. 
Among these reports, universal parameters such as ``reef areal dimensions, reef height, oyster density, and oyster size-frequency distribution'' are commonly assessed~\cite{beck2011oyster,pogoda2019current, blomberg2018habitat, mcfarland2018restoring, theuerkauf2019integrating}. 
The overall process is costly and labor-intensive, limiting the scalability of surveying efforts.

Similar challenges exist in on-bottom oyster aquaculture, where larval oysters are set on clutch and left to grow for years before harvest. Despite global growth for this economic sector~\cite{botta2020review}, the industry still relies on legacy equipment~\cite{kumar2018factors}. Advances in autonomy, which have improved terrestrial farming with real-time detection and precision techniques~\cite{yang2023real} could help underwater farming. By using these methods, farmers could better track oyster health, abundance, and size, optimize harvesting, and avoid areas with low yields or undersized oysters~\cite{stokesbury2011environmental}. 

In fact, by collecting videos or photographs of the beds and assessing the quantity of live and dead oysters, a census of a localized region can be produced to determine the quality of the reef and the harvestable biomass. Similar video-based methods were successfully used to assess habitat provisioning in oyster reefs and aquaculture gear to provide valuable estimates of relative abundance, species composition, and behavior~\cite{mercaldo2023oyster, veggerby2024shellfish, connolly2024estimating, armbruster11territorial}. However, these methods require highly trained individuals to manually identify and count animals over a large timespan, limiting the depth, scale, and applicability of monitoring for surveying. Autonomous systems can detect and assess live oysters in real time using vision models, offering a promising approach to monitoring. These systems enable on-site decision-making based on live feedback, but training such models requires large datasets of oyster imagery, which are often scarce. 

To address this, we leverage diffusion models to generate photorealistic synthetic images, augmenting real datasets.
These combined datasets are used to train models that are then deployed in autonomous underwater vehicles (AUVs) for real-time oyster population surveys.
The models trained on these combined datasets are deployed on AUVs, enabling autonomous surveys in oyster fields, and providing accurate counts of oyster populations within their biomes.
A key aspect of our approach is that the detection computation is performed in real-time onboard the AUV.

The contributions of this paper are as follows:
\begin{itemize}
\item The development and deployment of a state-of-the-art model for oyster detection with on edge implementation on the Aqua2 biomimetic AUV, to carry out tests on the oysters in real environment;
\item A novel scheme for the generation of photorealistic oyster imagery;
\item Open-sourcing the datasets and associated code to facilitate future research. 
\end{itemize}

The remainder of this paper is organized as follows. First, we place this work in the context of previous work in Sec.~\ref{section:related_work}. Then, we describe the pipeline used to create synthetic images and the details of the edge device within Sec.~\ref{section:problem_formulation}.
We then present extensive quantitative and qualitative evaluations of our approach in Sec.~\ref{section:Experiments_and_results}. Finally, we conclude the paper in Sec.~\ref{section:Conclusions} with parting thoughts on future work.

\section{RELATED WORK}
\label{section:related_work}
Robotic monitoring of aquatic environments presents technical challenges due to low visibility, complex terrains, and the need for autonomy and real-time decision-making~\cite{manjanna2016efficient,hansen2018autonomous, karapetyan2021meander}. 
Edge computing, as a framework that brings data processing applications closer to data sources, can support robot autonomy in environmental monitoring missions\cite{roostaei2023iot,ModasshirCRV2018}. 
When implemented in autonomous underwater vehicles (AUVs), edge computing allows real-time and \emph{local} data processing in support of a range of robotic missions, from object detection~\cite{abdullah2024caveseg,ModasshirRobio2018} to autonomous navigation~\cite{xanthidis2020navigation} and exploration~\cite{lin2024uivnav} under resource constraints. 

Within marine environmental monitoring, oyster detection, in particular, presents unique additional challenges due to the heavily occluded marine environment these creatures inhabit, which makes accurate manual labeling especially difficult.
Sadrfaridpour et al.~\cite{sadrfaridpour2021detecting} used Mask-R-CNN~\cite{he2017mask} for oyster detection; however, their dataset was small and collected on an oyster farm, which lacked the variety of a wild reef setting ---needed for robust detection in real-world environments. 
The collection and annotation of sufficient marine data, essential for achieving satisfactory results with vision models, has always been a challenge.

To save time and resources associated with collecting real data, we follow the conceptual approach proposed by Lin et al.~\cite{lin2023oysternet}, which uses 3D models to generate synthetic data. This technique has been successfully applied in various object detection tasks, such as whale detection with synthetic satellite images~\cite{gaur2023whale}, aerial object detection~\cite{lin2023seadronesim}, real and virtual segmentation~\cite{wu2024marvis}, and olive detection~\cite{karabatis2023detecting}. We improved upon OysterNet~\cite{lin2023oysternet} to generate realistic oyster assemblages and using stable diffusion for sim-to-real transfer. 

\begin{figure*}[ht!]
\vspace{3mm}
\includegraphics[width=0.95\textwidth]{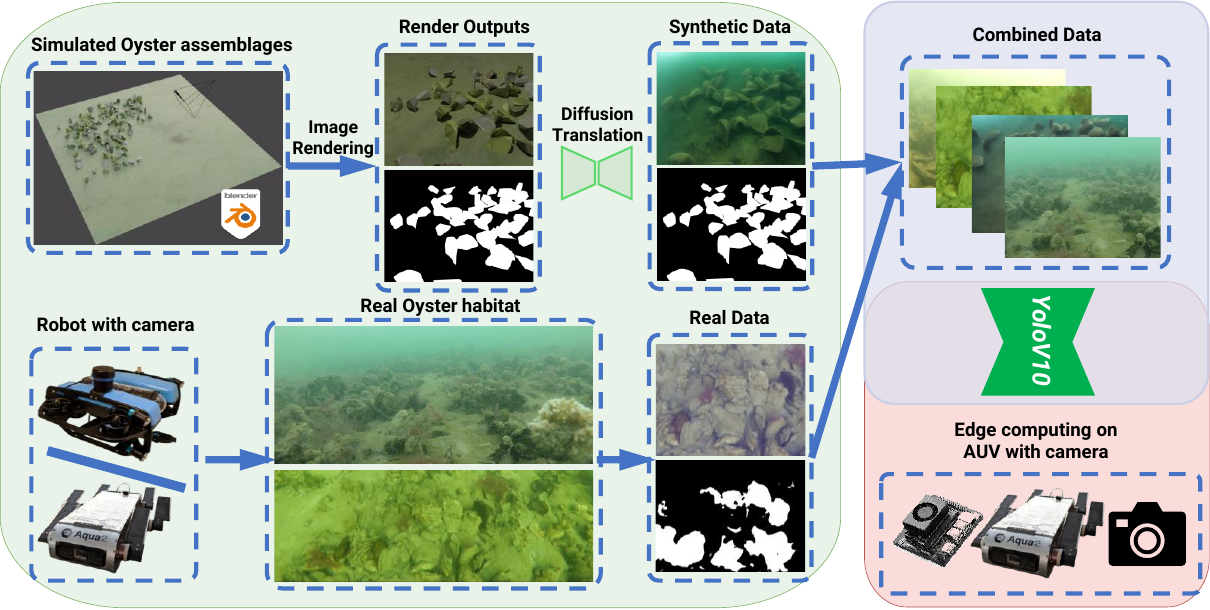}
\centering
\captionsetup{font={footnotesize},labelfont=bf}
\caption{Overview of the oyster detection system. Simulated oyster assemblages are rendered using Blender and paired with ground-truth segmentation masks. Diffusion models are employed to generate synthetic oyster imagery, ensuring consistency with real oyster habitats. Real-world data are collected by the Aqua2 
and BlueROV, equipped with a camera for capturing oyster images in the field. The combined dataset is used to train a YOLOv10, which is deployed on an edge computing platform on Aqua2 for real-time oyster detection.}
\label{fig:synth_data_pipeline}
\end{figure*} 

While Generative Adversarial Networks (GANs) have been used for sim-to-real tasks such as underwater image restoration~\cite{li2017watergan}, synthetic oyster generation~\cite{lin2023oysternet}, and AUV pose estimation~\cite{joshi2020deepurl}, diffusion models provide an attractive and often advantageous alternative. 
Indeed, diffusion models~\cite{ho2020denoising}, which generate images by refining random inputs, have proven effective for high-quality image synthesis~\cite{ghiasi2017exploring,jackson2019style,rombach2022high}. 
Recent advances, such as ControlNet~\cite{zhang2023adding}, enable better control over the output by integrating external conditions like depth and segmentation map. 
This added control not only improves the fidelity of the generated images, but also expands the utility of diffusion models, enabling them to be used for tasks like synthetic data augmentation and domain-specific image generation. 

Given the capability of these models to generate representative synthetic data, many have used diffusion models to aid in the training of vision models. He et al.~\cite{he2022synthetic} demonstrated the use of synthetic data in training vision models, stating its efficacy in improving the performance of models in zero-shot and low-shot environments. Bansal et al.~\cite{bansal2023leavingrealityimaginationrobust}
used synthetic data to achieve higher accuracy with ImageNet models, and Trabucco et al.  
Off-the-shelf stable diffusion models have also been used~\cite{Trabucco2023EffectiveDA} to enhance existing data and apply it to few-shot domains. Despite the obvious advantages that diffusion models afford for dataset enhancement, few efforts attempt to create stable diffusion-enhanced datasets for the \emph{underwater domain}.

Our work combines
(i) edge computing for real-time underwater exploration and
(ii) advanced synthetic data generation.
While previous efforts faced challenges due to small and homogeneous datasets, our approach leverages diffusion models and geometric modeling to create diverse and realistic oyster datasets and deploy state-of-the-art machine learning models such as YOLOv10~\cite{wang2024yolov10realtimeendtoendobject} running locally on the Aqua2 AUV, ensuring robust performance with edge computation in real-world scenarios. The resulting pipeline is crucial to developing robust robust oyster detection systems and improving the effectiveness of autonomous, real-time monitoring of complex marine environments. 

\section{System Overview}
\label{section:problem_formulation}
To efficiently monitor oyster habitats in real time, we utilize the Aqua2 robotic platform (Figs~\ref{fig:detection_result}, \ref{fig:experiment_drone}), designed for nondisruptive environmental monitoring tasks. We will first provide an overview of the Aqua2 robot, followed by a detailed explanation of the software deployed on it. To further address the challenge of limited real-world oyster imagery in reef environments, we supplement our dataset with synthetic data, which will be discussed in the section~\ref{Synthetic_Image_Generation} of this overview.

\subsection{Hardware Overview}
\label{Hardware_overview}
Aqua2 is a biomimetic hexapedal Autonomous Underwater Vehicle (AUV), designed for minimal environmental disturbance in sensitive habitats such as coral~\cite{ModasshirICRA2020,modasshir2021autonomous} and oyster reefs~\cite{dudek2005visually}. Unlike traditional AUVs that use thrust-based propulsion, Aqua2 employs a unique biomimetic reciprocating paddle motion with its six fins, significantly reducing sediment disruption by avoiding thrust wash. This feature is crucial to maintain water clarity in underwater environments, as sediment plumes can obscure sensors and cameras~\cite{joshi2022underwater}. The Aqua2 vehicle is equipped with a NVIDIA Jetson Xavier NX processor, which features a 384-core NVIDIA Volta GPU with 48 Tensor Cores, delivering up to 21 TOPS (Tera Operations Per Second) of performance at 15W of power. The processor is complemented by a 6-core NVIDIA Carmel ARM®v8.2 64-bit CPU, providing robust processing capabilities essential for complex, real-time image processing tasks. In addition, the vehicle is equipped with three 2-megapixel cameras optimized for low-light underwater imaging and with scene-based white balancing. The two front E-CAMs capture full HD video at 65 fps, while the rear Blue Robotics camera records at 30 fps. Both have focal lengths of about 3mm, but the front cameras offer a much wider field of view and overall better performance for surveying underwater environments.
On a full load, Aqua2 is capable of running for 5 hours in real world environments.

\subsection{Software Overview}
Aqua2's software architecture leverages the Docker containers for easy management and deployment.  NVIDIA’s Container Runtime, compatible with the Open Containers Initiative (OCI), provides GPU access as outlined in~\ref{Hardware_overview}.
Based on Dustynv/L4t-ml v-r36.2.0\footnote{\url{https://hub.docker.com/r/dustynv/l4t-ml}} container, we integrate Robot Operating System (ROS2), OpenCV, PyTorch {v2.2.0} and Ultralytics. The system runs on ROS2 Humble middleware for real-time communication capabilities, modular design, and simplicity. OpenCV handles basic computer vision tasks and Ultralytics installs YOLOv10 and all its dependencies. 
We developed a custom ROS package that executes the YOLOv10 ROS node to establish the vision pipeline. 
During operation, Aqua2’s vision pipeline captures images from a downward-facing camera. Pre-trained YOLOv10 submodels, cached on the robot, are quickly loaded to process and compare predictions, highlighting regions of interest and publish these images on the ROS network. These results can be analyzed in real-time by the operator for decision-making or used for autonomous monitoring tasks.
With the hardware and software infrastructure in place, we now turn our attention to addressing the challenge of limited real-world oyster imagery by leveraging synthetic data generation to further enhance the model’s performance.

\subsection{Synthetic Image Generation}
\label{Synthetic_Image_Generation}
Given the challenges of acquiring real-world data, synthetic image generation has become a valuable tool for extending datasets.
By using diffusion models with ControlNet, we transform a synthetic dataset, comprising segmentation masks and depth images generated in Blender, into realistic oyster imagery that closely resembles the target underwater environment, enhancing vision model performance. 
This method effectively bridges the gap between synthetic and real-world data, producing more representative images while retaining the flexibility of synthetic generation.
The data generation pipeline in Fig.~\ref{fig:synth_data_pipeline} illustrates the ControlNet inputs on the left side, consisting of target domain images, the synthetic blender images, depth images and segmentation ground truth masks rendered with blender.
For further guiadance a positive and negative prompt is provided. The details of the synthetic data generation pipeline are described in the following sections.

\begin{figure}[t!]
\vspace{5mm}
\includegraphics[width=\linewidth]{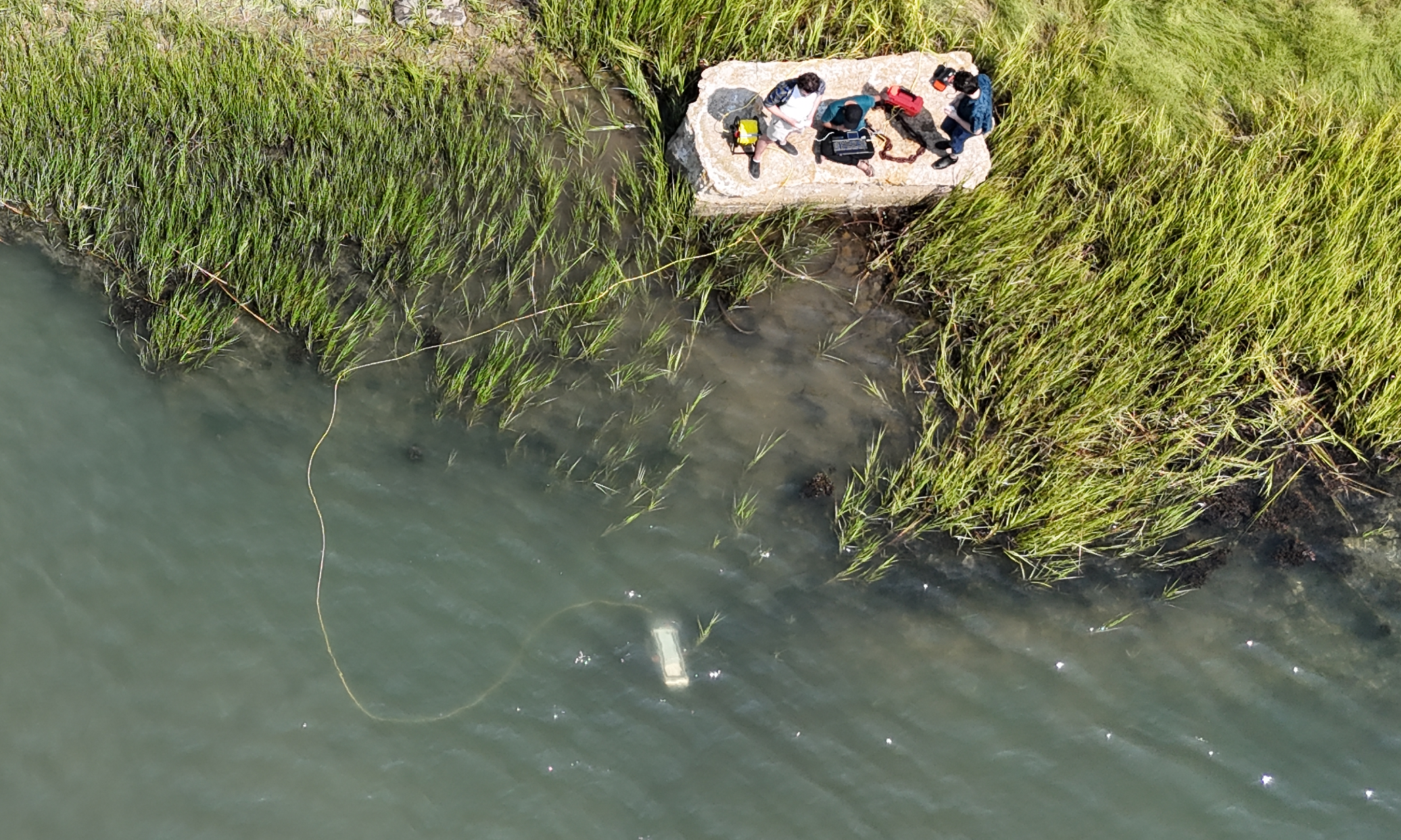}
\centering
\captionsetup{font={footnotesize},labelfont=bf}
\caption{Experimental deployment of the system, surveying a wild near-shore oyster reef in Lewes, DE. Operators can be seen teleoperating the Aqua2 and analyzing the feedback.}
\label{fig:experiment_drone}
\vspace{-4mm}
\end{figure}
\subsubsection{Simulation Image Rendering} 
\label{sub:blender_sim}
To model the geometry of an oyster, we first 3D scanned ten oysters to build a mathematical model that approximates the shape in two stages: a 2D perimeter model and a stratified 3D model. 
\begin{equation}
	B_{i,0}(t) =
\begin{cases}
1         & \text{if } t_{i+1}\geq t\geq t_{i}\\
0,        & \text{otherwise},
\end{cases}
\label{eq:B_t_0}
\end{equation}

\begin{equation}
	B_{i,k}(t) = \tfrac{t-t_i}{t_{i+k}-t_i}B_{i,k-1}(t)+ \tfrac{t_{i+k+1}-t}{t_{i+k+1}-t_{i+1}}B_{i+1,k-1}(t).
	\label{eq:B_t_k}
\end{equation}
The 2D perimeter is represented using two cubic B-splines shows in~\eqref{eq:B_t_0} and~\eqref{eq:B_t_k} for the top and bottom shell halves. This model is then extended to 3D by adding depth changes across layers. High-frequency details are simplified for computational efficiency, with visual textures compensating for intricate surface variations. This approach provides a scalable way to generate realistic oyster models. More details of the modeling can be found in OysterNet~\cite{lin2023oysternet}

The 3D modeled oysters are then randomly distributed within the Blender\texttrademark\, 3D game engine.
The camera view was angled slightly to better emulate the point of view of an AUV observing the oysters in situ (Fig.~\ref{fig:detection_result}), providing a more accurate visual representation of the AUV monitoring the oyster habitat. 

The simulation is used to generate depth maps and the corresponding segmentation masks, which are used as the source domain in the synthetic dataset generation.
Providing geometric consistency in ControlNet through the depth map indicates the locations of individual oysters within the segmentation mask. 
\begin{figure}[t]
\vspace{3mm}
\includegraphics[width=\linewidth]{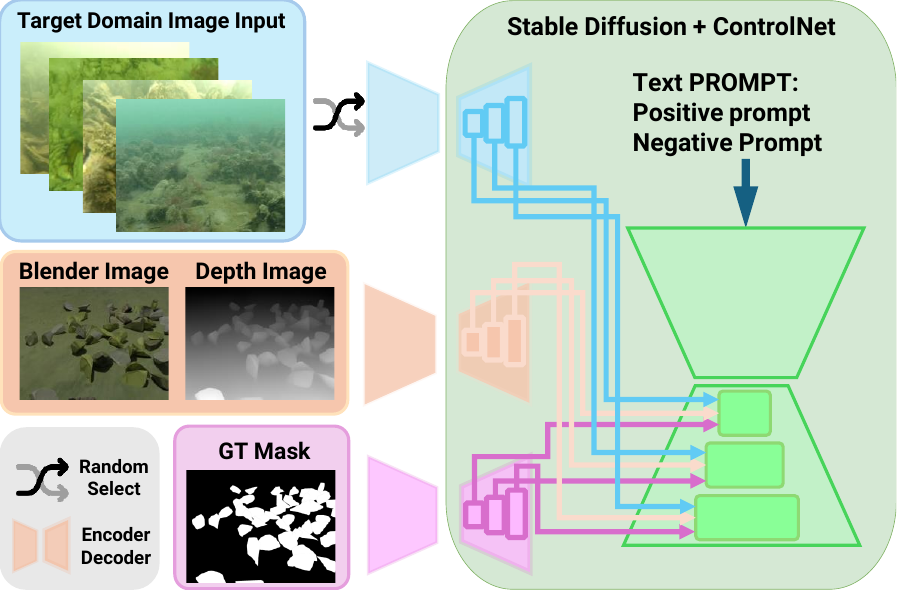}
\centering
\captionsetup{font={footnotesize},labelfont=bf}
\caption{Synthetic image generation pipeline using Stable Diffusion and ControlNet. ControlNet uses randomly sampled real underwater images combined with Blender-geneated images, depth images, and ground truth masks to ensure consistency. Stable Difussion model, guided by text prompts, and ControlNet, refines the synthetic data to closely match real-world oyster environments for vision model training.}
\label{fig:synth_data_pipeline}
\end{figure} 
\subsubsection{Real Data Collection}

To facilitate data collection, the BlueROV platform was equipped with a GoPro camera, and the Aqua2 robot was also employed to capture video of oysters in situ at Chesapeake Bay (Neavitt, MD; 38° 44' 41.0"N, 76° 18' 22.6"W) and Delaware Bay (Lewes, DE; 38°47'20.4"N 75°09'44.6"W) respctively. 
During the experiments, the robot was teleoperated to decouple the detection and classification results from the challenge of navigating around oyster structures, ensuring more precise data collection. 
From the video, high-quality frames were extracted and manually annotated by trained experts to identify the specific locations of the oyster within each frame.
This data was subsequently combined with the earlier OysterNet dataset to obtain the final combined, fully real dataset of more than 2000 manually annotated images. 
Furthermore, real images are used as ControlNet inputs that represent the target domain to generate synthetic images.

\subsubsection{Stable Diffusion and ControlNet}

Images generated as detailed in Section~\ref{sub:blender_sim} are provided as the source domain and iteratively selected as ControlNet inputs.
Depth maps and segmentation are kept in corresponding pairs such that the segmentation and geometry complement each other.
Therefore, guiding the image generation to align with the desired structure and spatial characteristics.
Additionally, four randomized reference images from the real data collection are used as image prompts for ControlNet to fine-tune the stylistic and environmental attributes of the generated images.
A carefully selected text prompt ensures that Stable Diffusion focuses on the specific target object, namely, the oysters, and it includes negative prompts to mitigate certain biases in the diffusion models. 
Each generated image from the diffusion model is paired with the corresponding segmentation mask and then used as one of the ControlNet inputs.
These masks serve to help generate annotations for the synthetic oyster images, enabling them to act as the ground truth.

Combining real and synthetic data ensures a sufficiently large dataset for training. With the system reviewed, we proceed to assess the performance of the system through a series of experiments.

\section{Experiments And Results}
\label{section:Experiments_and_results}

The mixed datasets obtained via the method described in ~\ref{Synthetic_Image_Generation} were then subsequently used to train all of the available YOLOv10 models and comparisons were run between these augmented models and the models which had only seen the real data. 

\subsection{Training Details}
To train the various models, 4000 oyster images were synthetically generated as described in Section~\ref{Synthetic_Image_Generation} with the size of 640$\times$480, which were then combined with the existing dataset of 2025 real images of 1920$\times$1680 resolution. To this end, we used only 30\% of the real dataset for training combined with all the synthetic dataset. We tested our method on the remaining 70\% of the real dataset to report our result.
We train all networks for 300 epochs with an initial learning rate of 0.01 with a cosine annealing learning rate scheduler. The Adam optimizer was used with momentum set to 0.937 and a weight decay of 0.0005. To avoid overfitting. The training batch size was 16, and the model was trained on resized images to 640$\times$640 pixels. In terms of data augmentation, we utilized several techniques, such as flipping, scaling, and random cropping. Close mosaic augmentation was applied for the first 10 epochs, with the mosaic probability set to 1.0. The model was further fine-tuned using automated augmentations, such as \textit{randaugment} and random erasing with a probability of 0.4.
During training, the object detection model was evaluated against a validation set, with an IoU threshold of 0.7. The maximum number of detections per image was capped at 300 to ensure efficient processing of the oyster habitat scenes.

\subsection{Experimental Results}

\renewcommand{\arraystretch}{2.5}   
\begin{table}[t!]
\vspace{2mm}
\captionsetup{font={footnotesize},labelfont=bf}
\caption{Quantitative Comparison of All YOLOv10 Sub-Model Performance}
\centering
\resizebox{\columnwidth}{!}{%
\begin{tabular}{cccccc}
  \hline
  Model & Inference Time (ms) & Frequency (Hz) & $AP_{50}$ & mAP50-95 \\
  \hline
  YOLOv10-N & 54.9 & \textbf{9.2} & 0.458 & 0.234 \\
  YOLOv10-S & 71.0 & 8.9 & 0.557 & 0.314 \\
  YOLOv10-M& 158.1 & 4.8 & 0.642 & 0.468 \\
  YOLOv10-B & 214.8 & 3.8 & 0.645 & 0.416 \\
  YOLOv10-L & 264.1 & 3.2 & \textbf{0.657} & 0.430 \\
  YOLOv10-X & 375.2 & 2.3 & 0.642 & 0.424 \\
  \hline
\end{tabular}%
}
\label{tab:yolov10}
\end{table}


Extensive testing of various YOLOv10 submodels was carried out at the Marine Operation boat basin facility of the University of Delaware in Lewes, Delaware Fig.~\ref{fig:experiment_drone}. This test focused on assessing the performance, accuracy and adaptability of these submodels in real-world scenarios, ensuring their robustness in handling the specific environmental and operational challenges presented by the location. 

Lighter models like Yolov10-N achieved faster inference times and higher frequencies, while more complex models like Yolov10-X were slower. For example, Yolov10-N processed a frame in 55 ms with a frequency of 9.2Hz (Table~\ref{tab:yolov10}\textcolor{red}{a}), while the Yolov10-S model, although more accurate, had a frequency of 8.9Hz with a processing time of 71 ms (Table~\ref{tab:yolov10}\textcolor{red}{b}).
\vspace{3mm}
\begin{figure}[t!]
\includegraphics[width=0.95\linewidth]{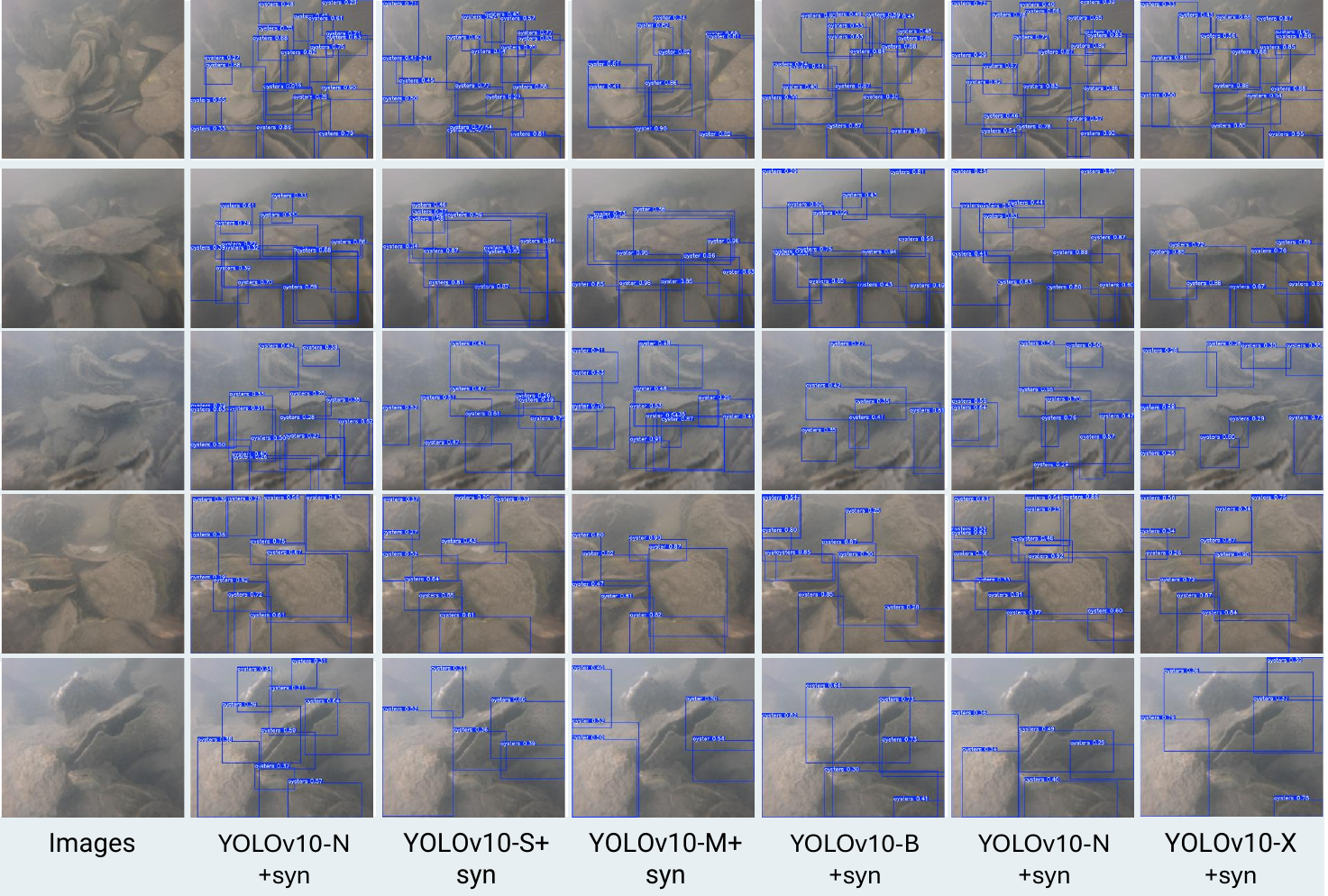}
\centering
\captionsetup{font={footnotesize},labelfont=bf}
\caption{Qualatative Comparison of All YOLOv10 Sub-Model Performance. \textit{This image is best seen in color on a computer screen at 400\% zoom.}}
\vspace{-3mm}
\label{fig:qualatative_result_yolov10_all}
\end{figure}

\begin{figure}[t!]
\vspace{3mm}
\includegraphics[width=0.95\linewidth]{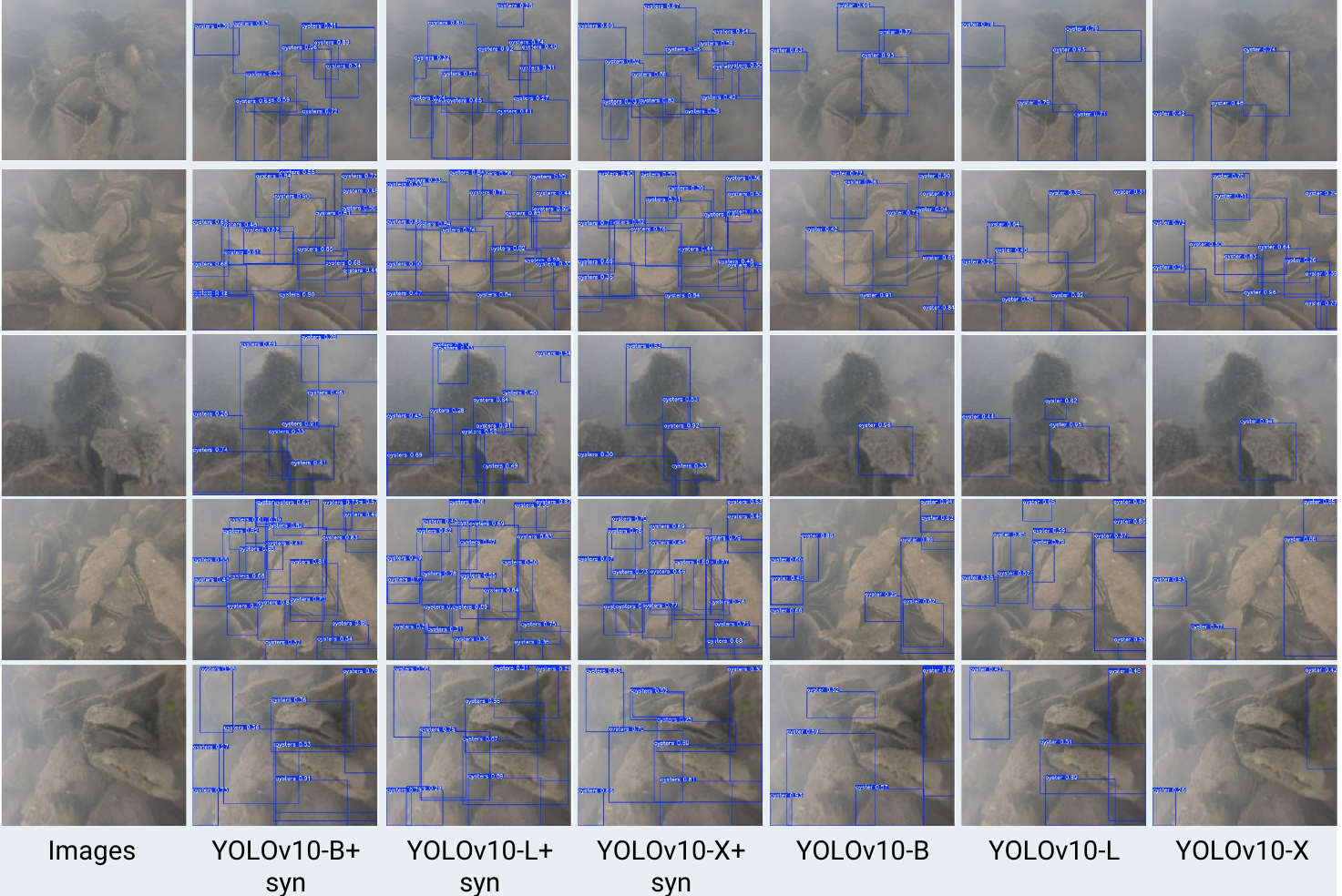}
\centering
\captionsetup{font={footnotesize},labelfont=bf}
\caption{Qualatative Comparison of YOLOv10-B, YOLOv10-L, and YOLOv10-X Models with and without Synthetic Data. \textit{This image is best seen in color on a computer screen at 400\% zoom.}}
\label{fig:qualatative_result_yolov10_syn}
\end{figure}

The YOLOv10-M model, suitable for general detection, processed frames in 158 ms at 4.8Hz (Table~\ref{tab:yolov10}\textcolor{red}{c}). YOLOv10-B offered a good compromise, with a processing time of 214 ms and a frequency of 3.8Hz (Table~\ref{tab:yolov10}\textcolor{red}{d}). For more accuracy, YOLOv10-L took 264 ms per frame at 3.2Hz (Table~\ref{tab:yolov10}\textcolor{red}{e}). Lastly, YOLOv10-X, the most computationally intensive model, was processed at 372 ms with a frequency of 2.3Hz (Table~\ref{tab:yolov10}\textcolor{red}{f}).

The qualitative results in Fig.~\ref{fig:qualatative_result_yolov10_all} from testing multiple YOLOv10 sub-models demonstrate a clear trend: generally, larger models achieve better detection accuracy, as indicated by higher mAP50 and mAP50-95 values. As depicted in Table~\ref{tab:yolov10}, YOLOv10-L outperforms smaller models like YOLOv10-N, with mAP50 scores of 0.657 compared to 0.458. However, this trend does not hold for the largest model, YOLOv10-X, where the performance slightly drops, suggesting that the size of the dataset might not be sufficient to fully leverage the complexity of the larger network. This indicates a point of diminishing returns with increasing model size, where the added complexity does not necessarily translate into better detection performance.

When comparing models trained with and without synthetic data in Fig.~\ref{fig:qualatative_result_yolov10_syn}, it is evident that incorporating synthetic data improves the overall detection results across the board. As depicted in Table~\ref{tab:yolov10_with_syn}, YOLOv10-L with synthetic data achieves an mAP50 of 0.657 compared to 0.638 without synthetic data. However, this improvement comes with a trade-off: the confidence in these detections, as reflected in the mAP50-95 scores, is slightly lower with synthetic data. This suggests that while synthetic data can enhance detection capabilities, it might introduce variability or noise that affects the precision of the model's predictions.


\begin{table}[t!]
\vspace{4mm}
\captionsetup{font={footnotesize},labelfont=bf}
\caption{Comparison of YOLOv10-B, YOLOv10-L, and YOLOv10-X Models with(S) and without(R) Synthetic Data. (S): the model is trained on combined synthetic and real data. (R): the model is trained on real data only}
\centering
\resizebox{\columnwidth}{!}{%
{\huge
\begin{tabular}{ccc|ccc}
  \hline
  Model (S) & $AP_{50}$ & mAP50-95 & Model (R) & mAP50  & mAP50-95 \\
  \hline
  YOLOv10-B & 0.645 & 0.416 & YOLOv10-B & 0.639 & 0.472 \\
  \hline
  YOLOv10-L & \textbf{0.657} & 0.430 & YOLOv10-L & 0.638 & 0.478 \\
  \hline
  YOLOv10-X & 0.642 & 0.424 & YOLOv10-X & 0.630 & \textbf{0.480} \\
  \hline
\end{tabular}%
\vspace{-8mm}
}
}
\label{tab:yolov10_with_syn}
\end{table}

\section{Conclusion and Future Work}
\label{section:Conclusions}
In this work, we present a comprehensive system for real-time, on-edge, underwater oyster monitoring, which integrates advanced object detection models and synthetic data generation via stable diffusion on the Aqua2 underwater robotic platform. We demonstrated that systematically combining robotic capabilities with domain-specific synthetic data enhances model accuracy and detection performance for underwater environments. This system not only improves oyster monitoring efficiency but also establishes a framework for broader applications in marine environmental monitoring.

Looking ahead, the system can be improved both from hardware and software perspectives. On the hardware side, we could introduce additional sensor modalities, such as multi-beam ultrasound and single-beam ultrasound, for depth estimation, which would significantly enhance performance in murky water conditions. From a software standpoint, incorporating advanced navigation and exploration algorithms would enable the robot to autonomously navigate and perform surveillance tasks. Online learning algorithms could allow the system to adapt in real-time to different environmental conditions, such as variations in water color. Finally, introducing collaborative multi-robot systems would make monitoring more efficient and comprehensive, allowing for larger-scale underwater environmental monitoring.



\section*{ACKNOWLEDGMENT}

This work was supported by USDA NIFA Sustainable Agriculture System Program under award number 20206801231805, NOAA's Project ABLE via award number NA22OAR4690620-T1-01, the NSF awards 1943205 and  2024741, the EU-program EC Horizon 2020 for Research and Innovation under grant agreement No. 101120823, project MANiBOT, and the US Army Corps of Engineers, ERDC Contracting Office under Contract No. W912HZ-22-2-0015.
We thank Dr. Christopher Rasmussen and Dr. Arthur Trembanis for their support during the 2024 Autonomous Systems Bootcamp at the University of Delaware, Independent Robotics for assisting with Aqua AUV operations, and Rileigh Hudock, Dr. Edward Hale, and Max Collins for help with data collection.
\bibliographystyle{IEEEtran}
\bibliography{IEEEabrv,refs}
\end{document}